\pdfoutput=1

\documentclass[11pt]{article}

\usepackage[final]{acl}

\usepackage{times}
\usepackage{latexsym}
\usepackage{ragged2e}
\usepackage{rotating,tabularx,siunitx}
\definecolor{cYellow}{RGB}{255,255,3}
\definecolor{cBlue}{RGB}{69,123,157}
\definecolor{cRed}{RGB}{231,56,71}
\definecolor{cRed_1}{RGB}{191,30,46}
\definecolor{cGray}{RGB}{168,218,219}
\definecolor{cBlue_2}{RGB}{5,48,97}
\definecolor{cBlue_1}{RGB}{115,186,214}
\definecolor{cBlue_3}{RGB}{13,76,109}
\definecolor{cBlue_4}{RGB}{64,121,160}
\definecolor{cOrange}{RGB}{250,134,0}
\definecolor{cBlue_6}{RGB}{13,76,109}
\definecolor{cBlue_7}{RGB}{16,106,130}
\definecolor{cBlue_8}{RGB}{19,136,160}
\usepackage{float}
\usepackage{graphicx}
\usepackage{subfigure}
\usepackage{caption}
\usepackage{amsmath,amssymb}
\usepackage{mathtools}
\usepackage{booktabs}
\usepackage{multirow}
\usepackage{multicol}
\usepackage{siunitx}
\usepackage{adjustbox}
\usepackage{pifont}
\usepackage{tabularx}
\usepackage{enumitem}
\usepackage{array, makecell, multirow}
\newcolumntype{P}[1]{>{\centering\arraybackslash}p{#1}}
\usepackage{siunitx}
\usepackage{soul}
\usepackage{xcolor}
\usepackage{colortbl}
\sethlcolor{cYellow}
\usepackage{tcolorbox}
\tcbuselibrary{skins, breakable, theorems}

\usepackage[T1]{fontenc}

\usepackage[utf8]{inputenc}

\usepackage{microtype}

\definecolor{colorxmark}{RGB}{255, 87, 51}
\definecolor{colorcmark}{RGB}{66, 154, 137}
\definecolor{headcolor}{HTML}{018161}
\definecolor{relationcolor}{HTML}{d95f02}
\definecolor{tailcolor}{HTML}{6560a3}
\definecolor{fkscp_color}{HTML}{ff8400}
\definecolor{mind_color}{HTML}{943dff}
\definecolor{new_knowledge_color}{HTML}{7030A0}
\definecolor{original_tail_color}{HTML}{BF9000}

\usepackage{inconsolata}
\newcommand*{\affaddr}[1]{#1} 
\newcommand*{\affmark}[1][*]
{\textsuperscript{#1}}

\newcommand{\name}[1]{\textsc{Mind}} 
\newcommand{\taskone}[1]{\textsc{IntentionUnderstanding}}
\newcommand{\tasktwo}[1]{\textsc{IntentionUtilization}}

\title{\name{}: Multimodal Shopping Intention Distillation from Large Vision-language Models for E-commerce Purchase Understanding}

\author{
Baixuan Xu\affmark[1]\thanks{\quad Equal Contribution} ,
Weiqi Wang\affmark[1]$^{*}$,
Haochen Shi\affmark[1],
Wenxuan Ding\affmark[1],
Huihao Jing\affmark[1],\\
\textbf{
Tianqing Fang\affmark[1],
Jiaxin Bai\affmark[1],
Xin Liu\affmark[2],
Changlong Yu\affmark[2],
Zheng Li\affmark[2]},\\
\textbf{
Chen Luo\affmark[2],
Qingyu Yin\affmark[2],
Bing Yin\affmark[2],
Long Chen\affmark[1],
Yangqiu Song\affmark[1]}\\
\affaddr{\affmark[1]Department of Computer Science and Engineering, HKUST, Hong Kong SAR, China}\\
\affaddr{\affmark[2]Amazon.com Inc, Palo Alto, CA, USA} \\
\texttt{bxuan@connect.ust.hk, \{wwangbw, longchen, yqsong\}@cse.ust.hk}\\
\texttt{\{xliucr, changlyu, amzzhe, cheluo, qingyy, alexbyin\}@amazon.com}\\
}

\begin{document}
\maketitle
\begin{abstract}
Improving user experience and providing personalized search results in E-commerce services heavily rely on understanding purchase intention. 
However, existing methods for acquiring large-scale intentions bank on distilling large language models with human annotation for verification. 
Such an approach tends to generate product-centric intentions, overlook valuable visual information from product images, and incurs high costs for scalability.
To address these issues, we introduce~\name{}, a multimodal framework that allows Large Vision-Language Models (LVLMs) to infer purchase intentions from multimodal product metadata and prioritize human-centric ones. 
Using Amazon Review data, we apply~\name{} and create a multimodal intention knowledge base, which contains 1,264,441 intentions derived from 126,142 co-buy shopping records across 107,215 products. 
Extensive human evaluations demonstrate the high plausibility and typicality of our obtained intentions and validate the effectiveness of our distillation framework and filtering mechanism. 
Further experiments reveal the positive downstream benefits that \name{} brings to intention comprehension tasks and highlight the importance of multimodal generation and role-aware filtering. Additionally, \name{} shows robustness to different prompts and superior generation quality compared to previous methods.
Our code and data are publicly available at \href{https://github.com/HKUST-KnowComp/MIND_Distillation}{https://github.com/HKUST-KnowComp/MIND\_Distillation}.
\end{abstract}

\begin{figure}[t]
    \centering
    \includegraphics[width=1\linewidth]{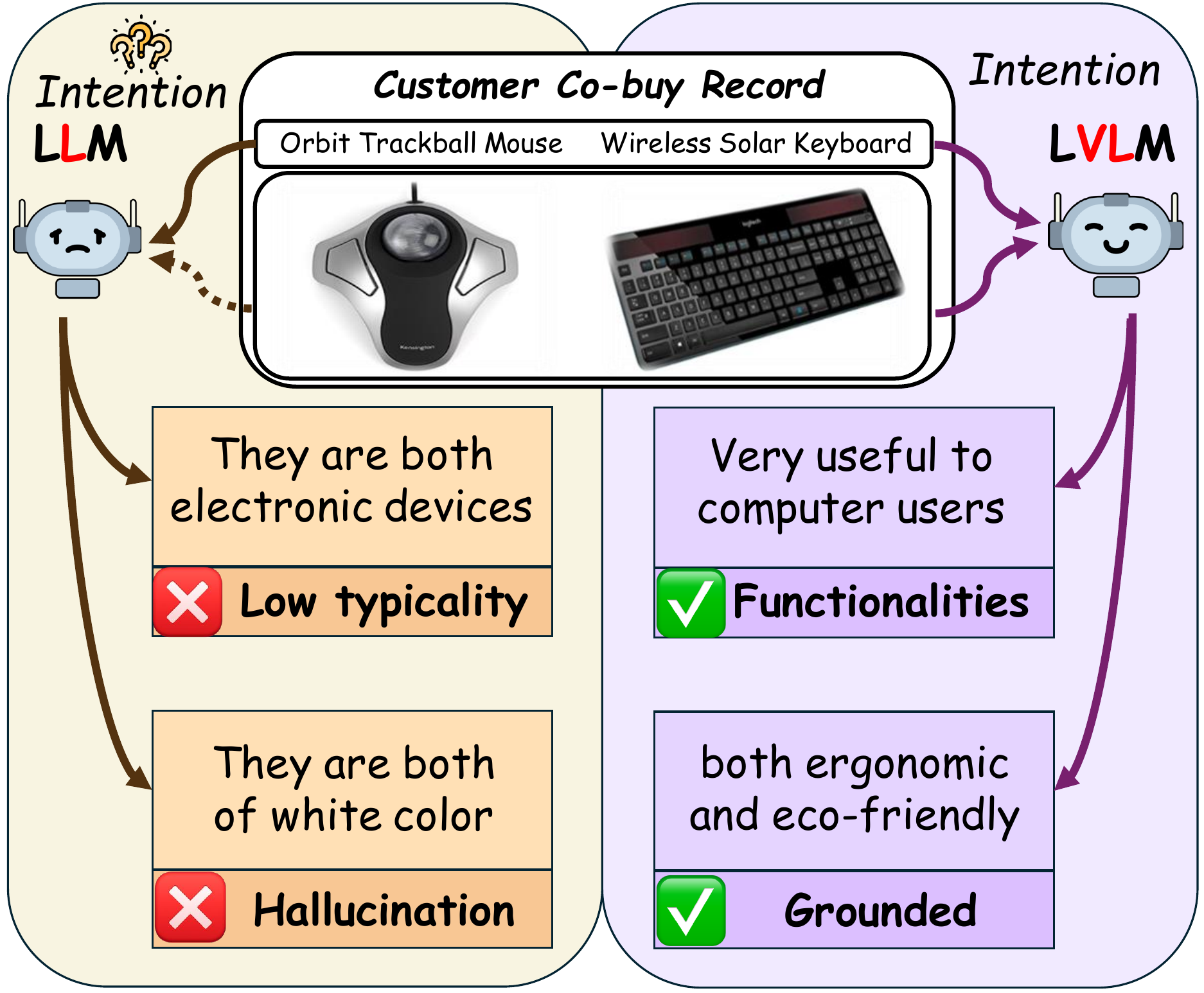}
    \vspace{-0.1in}
    \caption{
    Examples showing the process of distilling purchase intentions from \textcolor{fkscp_color}{large language models} and \textcolor{mind_color}{large vision-language models}.
    Without product images, \textcolor{fkscp_color}{large language models} tend to generate intentions with low typicality and hallucinated facts, while leveraging \textcolor{mind_color}{large vision-language models} resolve such issue.
    }
    \label{fig:intro}
    \vspace{-0.2in}
\end{figure}

\section{Introduction}
Understanding customers' intentions behind their purchase behaviors remains crucial in E-commerce as it potentially benefits several downstream tasks, such as product recommendation~\cite{DBLP:conf/kdd/GrbovicRDBSBS15,DBLP:conf/kdd/ZhaoGHJWL14,DBLP:conf/cikm/LiZC20} and search query answering~\cite{DBLP:conf/cikm/ZhaoCY19,DBLP:conf/sigir/HirschGNDK20,DBLP:conf/nips/BaiLW0S23}.
Unlike traditional factual knowledge related to products, intentions are implicit mental states of customers, which typically require commonsense knowledge to understand and reason upon~\cite{bratman1984two}.
For example, in Figure~\ref{fig:intro}, the intentions of purchasing a mouse and a keyboard can be \textit{they are very useful to computer users}, which is not mentioned either in the customer's query or products' metadata.
Thus, due to such implicitness, it is infeasible to perform large-scale automatic extraction from text to obtain them.


To combat this,~\citet{DBLP:conf/acl/YuWLBSLG0Y23} proposed to distill purchase intentions from large language models, such as OPT~\cite{OPT}, by prompting them with real purchasing records and relevant product metadata.
Human-in-the-loop annotations are also carried out to verify the plausibility and typicality of the generated intentions and train a discriminator for large-scale critic filtering.
\citet{yu2024cosmo} further entangled human annotations with instruction tuning to align the distilled intentions with a human-centric perspective.
While these works provide a straightforward approach to intention acquisition, several limitations still persist.

First, previous works on E-commerce intention knowledge base construction have solely focused on the text modality, thereby sacrificing significant supervision signals from visual modalities, such as product images. 
This oversight hinders the model from obtaining a more comprehensive understanding of the product, consequently compromising the quality of the generated intentions, as demonstrated in the left lower part of Figure~\ref{fig:intro}.
Furthermore, recent work has shown that intentions derived using current distillation methods exhibit bias towards product-centric aspects, excessively emphasizing product properties and metadata~\cite{DBLP:journals/corr/abs-2402-14901}. 
Consequently, interactions between the products and customers, including potential use cases and features of interest to customers, are absent from the derived intentions, despite being fundamental in facilitating customers' shopping experience.
Finally, human annotations are heavily deployed in current intention collection methods, which serve as a critical step in controlling the quality of the generated results. 
This poses a challenge towards constructing scalable yet diverse intention knowledge bases with minimum human supervision cost.

To address these issues, we propose~\name{}, a \textbf{\underline{M}}ultimodal Shopping \textbf{\underline{I}}ntentio\textbf{\underline{N}} \textbf{\underline{D}}istillation framework. 
\name{} instructs Large Vision-Language Models (LVLMs) to generate purchase intentions in a three-step manner, based on real user co-buy records and product metadata. 
Specifically, we select LLaVa~\cite{DBLP:conf/nips/LiuLWL23a} as a representative LVLM and incorporate both visual information from the product images and text information from the product name into the generation process.
To better align the generated raw intentions with human preferences and alleviate human annotation costs for further quality control, we propose a human-centric role-aware mechanism. 
This mechanism first instructs LLaVa to discover similar features between the products and then imitates a customer agent to decide whether the products would be bought together under previously generated intentions.

By applying~\name{} to a subset of the Amazon Review Dataset~\cite{DBLP:conf/emnlp/NiLM19}, we construct a multimodal intention knowledge base. 
It features 1.26 million of intentions over 126,142 co-buy shopping records across 107,215 products. 
Human evaluations further confirm: (1) the exceptional quality of our generated intentions, which have higher plausibility and typicality than previous generation methods, and (2) the effectiveness of our proposed human-centric role-aware mechanism.
Furthermore, we apply our generated intentions to two downstream tasks in the IntentionQA benchmark~\cite{ding2024INTENTIONQA}, which evaluates a language model's abilities to discriminate and utilize purchase intentions. 
Extensive experiments show that distilling our generated intentions into large language models' provide substantial benefits on both tasks via fine-tuning.
Further ablation studies reveal the importance of incorporating visual cues of products in~\name{} and the necessity of integrating our proposed role-aware filter mechanism. 
Moreover, analyses demonstrate the remarkable diversity of \name{}'s intentions and the exceptional robustness of~\name{} when using different prompts.

\section{Related Works}
\subsection{Shopping Intention in E-commerce}
Shopping intention is an implicit mental state that motivates purchase-related behaviors from the customer's perspective~\cite{koo2010interactional}. 
Various studies have been conducted to examine the impact of consumer shopping intentions on downstream applications~\cite{DBLP:conf/www/DaiZNWWL06,DBLP:conf/www/ZhangFDY16,DBLP:conf/wsdm/HaoHPWYWW22,EcomScript}. 
Recently,~\citet{DBLP:conf/emnlp/NiLM19} suggested using customer reviews to investigate the underlying purchase intentions in consumer purchase behavior and created a large-scale review dataset based on Amazon. 
Building upon this,~\citet{DBLP:conf/acl/YuWLBSLG0Y23} proposed FolkScope, which aims to guide LLMs in generating user co-buy intentions for different product pairs by grounding them in ConceptNet relations~\cite{DBLP:conf/aaai/SpeerCH17}. 
While human evaluations confirmed its effectiveness,~\citet{DBLP:journals/corr/abs-2402-14901} argued that it not only remains expensive to scale up but also fails to align the resulting shopping intentions with human preferences, which encompass a wide range of factors beyond product properties and similarities. 
To tackle these issues, in our work, we propose~\name{}, a framework that undermines online co-buy intentions and aligns better with human perceptions.

\begin{figure*}
    \centering
    \includegraphics[width=1\linewidth]{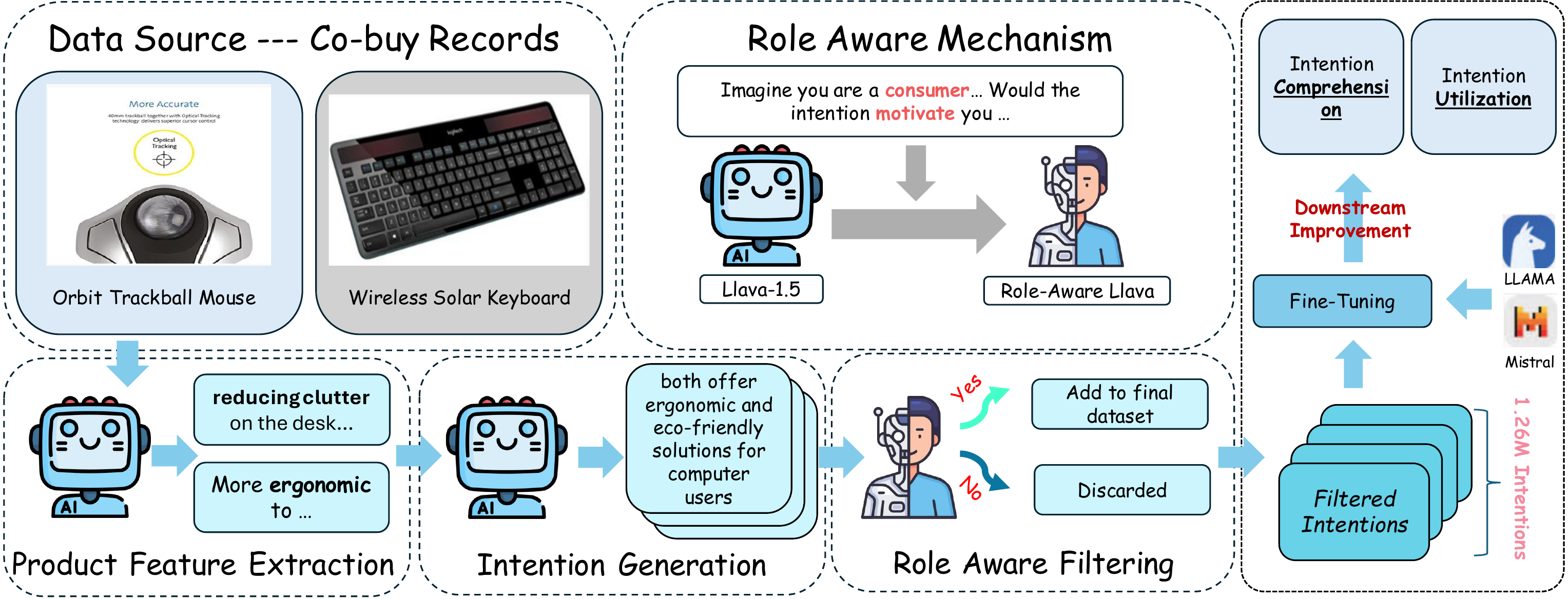}
    \vspace{-0.25in}
    \caption{An overview of \name{}.
    We first extract features from products in real-world co-buy records, generate intentions multimodally, and apply a human-centric role-aware filter for quality optimization.}
    \label{fig:framework_overview}
    \vspace{-0.25in}
\end{figure*}

\subsection{Multimodal Knowledge Distillation}
Since VLMs have yield significant advance recently~\cite{DBLP:conf/nips/LiuLGY0WYS23,DBLP:conf/icml/0008LSH23,DBLP:journals/corr/abs-2304-10592, DBLP:conf/eacl/ChanCWJFLS24, DBLP:conf/argmining/ZongWXZSWSWS23}, distilling domain-specific knowledge from them has become an effective yet cost-friendly trend in multimodal studies~\cite{DBLP:conf/wsdm/LiuLSYZ23,DBLP:journals/corr/abs-2402-18169,DBLP:conf/emnlp/JinSNT0F21}. 
\citet{DBLP:conf/wsdm/LiuLSYZ23} proposed a framework that applies self-distillation to stimulate the pre-train process of BERT to improve its performance in E-commerce product understanding tasks.
\citet{DBLP:journals/corr/abs-2402-18169} similarly instructed MiniGPT4~\cite{DBLP:journals/corr/abs-2304-10592} to generate user intention form social media posts text and its associated images.
\citet{DBLP:conf/emnlp/JinSNT0F21} also designed a framework to instruct the student model to imitate teacher model's behavior, which successfully preserved the teacher model's capabilities with fewer parameters.
In our work, we share the same aspiration and leverage distillation as a tool for data collection that provides downstream benefits in the E-commerce domain.
Specifically, we designed a framework to distill E-commerce intentions from LLaVa~\cite{DBLP:conf/nips/LiuLWL23a} and construct a comprehensive intention knowledge base based on the resulted generations. 

\section{The \name{} Framework}
\subsection{Overview of \name{}}
Following~\citet{DBLP:conf/acl/YuWLBSLG0Y23}, the objective of~\name{} is formulated as a text generation task. 
Given a record that shows a customer's co-buy (purchasing together) of two products, along with the detailed metadata of both products,~\name{} aims to generate the intentions behind such purchase behaviors that best align with the customer's mental state during the purchase, which includes their beliefs, desires, and intents~\cite{georgeff1999belief}.

Formally, for a given co-buy record, we define the two products as $p^1$ and $p^2$, along with their associated images $\{p^1_i, p^2_i\}$, and features and attributes $\{p^1_f,p^2_f\}$.
\name{} aims to leverage a LVLM $F$ to generate the intentions $I(p^1,p^2)$ of purchasing both products based on a pre-defined commonsense relation $r$, denoted as $I(p^1,p^2,r)=F(p^1_f,p^2_f,p^1_i, p^2_i,r)$.
In this paper, we follow~\citet{DBLP:conf/acl/YuWLBSLG0Y23,yu2024cosmo} and use relations from ConceptNet~\cite{DBLP:conf/aaai/SpeerCH17} to model the intentions. 
LLaVa-1.5-13b~\cite{DBLP:conf/nips/LiuLWL23a} is used as the LVLM $F$.

To achieve this objective, we design three sequentially connected steps within~\name{}, which are shown in Figure~\ref{fig:framework_overview}.
These steps are termed as: 
(1) product feature extraction; 
(2) co-buy intention generation; 
(3) human-centric role-aware filtering. 
Together, they form a collective pipeline for systematic intention acquisition without the need for human supervision and quality filtering.



\subsection{Source Data Collection}
We utilize the Amazon Review Data released by~\citet{DBLP:conf/emnlp/NiLM19}, which contains millions of products from 18 domains. 
Each product is accompanied by detailed reviews, co-buy records, and metadata, including its product title, features, attributes, and images provided by the retailer. 
Following~\citet{DBLP:conf/acl/YuWLBSLG0Y23}, we select products from the \textit{Electronics} and \textit{Clothing, Shoes and Jewelry} domains as representative products to demonstrate the effectiveness of~\name{}. 
It is important to note that while this work focuses on only two domains, the \name{} framework itself is not limited to these.
It can be applied to any product domain, allowing for seamless deployment to undermine co-buy intentions across a wide range of products beyond those selected.
To fit our framework, we filter out products without accessible images that may have been removed from the Amazon service.

\subsection{Product Feature Extraction}
We begin processing the collected products by first extracting key features with the aid of LVLMs.
This is motivated by our observations that product descriptions and attributes, inputed by retailers, tend to be noisy and unorganized, probably for promotion and style organization purposes.
Thus, we explicitly instructs LVLMs to augment source product metadata by extracting implicit features from each product's image and title by leveraging a zero-shot prompt:
\begin{tcolorbox}[title={Prompt Template for Product Feature Extraction}, colback = cBlue_1!10, colframe = cBlue_7,  coltitle=white,fonttitle=\bfseries\small, center title,fontupper=\small,fontlower=\small]
\textbf{Visual Input:} $p_i$
\tcblower
\textbf{Textual Input:}
\texttt{<Instruction>}. Given the product shown in the image: $p_f$, generate additional features by focusing on the product's attribute, design, and quality.
\end{tcolorbox}
where \texttt{<Instruction>} is a detailed task instruction, and $p_i, p_f$ are the respective image and details (title, descriptions, etc.) of the product.
This enables LVLM to comprehend the product from both visual and textual modalities, thereby providing us with a richer set of features that complements those provided by the retailers.



\subsection{Co-buy Intention Generation}
Then, for each co-buy pair of products $(p^1,p^2)$, we provide LVLM with the acquired features together with all details of both products, and instructs it again to reason the intentions for purchasing them simultaneously.
Specifically, we follow~\citet{DBLP:conf/acl/YuWLBSLG0Y23} and leverage 20 commonsense relations from ConceptNet~\cite{DBLP:conf/aaai/SpeerCH17} as waymarks to lead LVLM in generating purchase intentions with controllable commonsense groundings.
Similar to the previous step, a zero-shot prompt is used:
\begin{tcolorbox}[title={Prompt Template for Intention Generation}, colback = cBlue_1!10, colframe = cBlue_7,  coltitle=white,fonttitle=\bfseries\small, center title,fontupper=\small,fontlower=\small]
\textbf{Visual Input:} $p^1_i, p^2_i$
\tcblower
\textbf{Textual Input:}
\texttt{<Instruction>}. A customer purchased a pair of products, as shown in the images. They are: $p^1_f, p^2_f$. Act as the customer and infer a potential intention behind such purchase. Start the intention with \texttt{<Relation>}.
\end{tcolorbox}
Where \texttt{<Instruction>} is a detailed task instruction and \texttt{<Relation>} is the corresponding text template of a commonsense relation from ConceptNet. 
For every relation, we generate only one intention per pair of products due to the large amount of products and co-buy records. 
However, this is not restricted and can easily scale up.





\subsection{Human-centric Role-aware Filtering}
To effectively manage a large amount of purchase intentions, quality control measures have become imperative.
While previous works relied on human annotaions for this purpose, recent works~\cite{DBLP:journals/corr/abs-2402-14901} show that co-buy intentions generated by LLMs, despite undergoing human filtering, still fail in capturing the customers' mental states but rather focus on factual similarities of the products, as demonstrated in Figure~\ref{fig:intro}.
This phenomenon, refered to as ``product-centric,'' restricts the potential downstream applications of the generated intentions.
To address both issues, inspired by recent works on theory-of-mind~\cite{DBLP:journals/corr/abs-2302-02083}, we propose to incorporate a filtering module, powered by a LVLM, after the generation process. 
We instruct the LVLM to assume the role of an E-commerce customer and provide it with a generated intention as the objective in the customer's mental state. 
Based on this intention, we present the LVLM with a pair of products and ask it to first determine whether the intention successfully motivates the purchase behavior and then generate a rationale to support its decision.
This process simulates a real-world scenario where the LVLM functions as a customer, making purchase decisions. 
By filtering intentions that result in a positive response for purchasing, we obtain intentions that are ``human-centric'' in the sense that they satisfy the mental state of an agent that is aware of its role as a customer. 
We term this approach as \textit{human-centric role-aware filtering}, which serves as an automatic filter to replace manual annotations. 
We apply this module to all the intentions we collected in previous steps and select the product-intention pairs that are accepted by the module as the final outcomes of our framework.
Detailed prompts are provided in Appendix~\ref{appendix:prompts_used}.

\section{Intrinsic Evaluations}
By applying~\name{} to products we collected from Amazon Reviews~\cite{DBLP:conf/emnlp/NiLM19}, we construct a multimodal intention knowledge base, with statistics shown in Table~\ref{tab:annotation_statistic}. 
In total, 1.26 million intentions are preserved after applying our proposed filtering module, spanning across 20 relations. 
Therefore, in this section, we first evaluate \name{} intrinsically by examining the quality of the generated intentions and the effectiveness of our proposed filter module through human annotation.

\subsection{Annotation Setup}
We hire human annotators from the Amazon Mechanical Turk service to evaluate the generated intentions.
For a generated intention, we task each worker to evaluate four aspects:

$\bullet$ \textbf{Plausibility} refers to the degree to which an intention of a co-buy purchase appears correct and reasonable given both products.

$\bullet$ \textbf{Typicality} evaluates how well the intention reflects a specific feature that causes the user behavior, which emphasizes on \textit{informativeness} and \textit{causality}~\cite{DBLP:conf/acl/YuWLBSLG0Y23}.

$\bullet$ \textbf{Human-centric} evaluates the extent to which the intention considers and aligns with the mental state and preferences of a human customer.

$\bullet$ \textbf{Filter rationale} evaluates the correctness of the reasoning or justification provided by the filtering module for accepting or rejecting a product-intention pair.

For each aspect, we ask the annotators to rate them as a binary classification task.
A random sample of 5,000 generated intentions are annotated, and the final vote is determined by the majority vote from three annotators.
The requirement for the annotators could be found in Appendix~\ref{appendix:annotator_requirement}.

\subsection{Annotation Results}
\label{sec:annotation_results}
The results of the annotations are presented in Table~\ref{tab:annotation_statistic}.
The annotators achieved a pairwise agreement of 73.1\% and a Fleiss's $\kappa$~\cite{fleiss1971measuring} of 0.56, indicating satisfactory internal agreement. 
The results reveal that~\name{} effectively generates purchase intentions that are both highly plausible (\textbf{94\%} on average) and typical (\textbf{90\%} on average) across all relations. 
This indicates the strong product understanding and intention reasoning capabilities of~\name{}. 
Additionally, our proposed human-centric role-aware filter correctly identifies 82\% of intentions on average, with 80\% of them having appropriate justifications for filtering.
These high percentages further validate the effectiveness of our proposed method, which serves as a cost-efficient and highly reliable quality control measure, replacing the need for human annotations. More details and analysis regarding the filtered out intentions are attached in Appendix~\ref{appendix:error_analysis_filter}

\begin{table}
\centering
\small
\begin{tabular}{l|l|llll}
\toprule
Relation&\#Int. &Pla.&Typ.&Fil.&Rat.\\
\midrule
Effect & 97,047 & 0.90 & 0.83 & 0.73 & 0.70  \\
MannerOf & 50,563 & 0.93 & 0.89 & 0.83 & 0.82\\
isA & 62,069 & 0.94 & 0.88 & 0.82 & 0.80 \\
Other & 545 & 0.94 & 0.90 & 0.79 & 0.75 \\
MadeOf & 40,593 & 0.95 & 0.92 & 0.85 & 0.82 \\
SimilarTo & 63,558 & 0.94 & 0.87 & 0.83 & 0.80 \\
UsedFor & 52,383 & 0.94 & 0.88 & 0.81 & 0.79 \\
Can & 90,392 & 0.95 & 0.91 & 0.82 & 0.78\\
CauseDesire & 95,097 & 0.94 & 0.90 & 0.82 & 0.80\\
RelatedTo & 64,152 & 0.93 & 0.89 & 0.81 & 0.79 \\
PartOf & 81,230 & 0.92 & 0.87 & 0.79 & 0.77 \\
Open & 122,296 & 0.93 & 0.89 & 0.83 & 0.82 \\
CreatedBy & 35,723 & 0.94 & 0.88 & 0.78 & 0.76 \\
DeriveFrom & 60,347 & 0.95 & 0.89 & 0.80 & 0.77 \\
DefinedAs & 51,680 & 0.96 & 0.92 & 0.84 & 0.84 \\
PropertyOf & 57,947 & 0.97 & 0.90 & 0.83 & 0.82 \\
CapableOf & 86,772 & 0.95 & 0.90 & 0.82 & 0.82\\
Cause & 61,860 & 0.95 & 0.92 & 0.83 & 0.82 \\
SymbolOf & 64,477 & 0.95 & 0.92 & 0.84 & 0.82 \\
DistinctFrom & 27,710 & 0.94 & 0.89 & 0.84 & 0.83\\
\midrule
Total & 1,264,441 & 0.94 & 0.90 & 0.82 & 0.80 \\
\bottomrule
\end{tabular}
\caption{Statistics of the intention knowledge base constructed via~\name{} and human annotation results.}
\vspace{-0.2in}
\label{tab:annotation_statistic}
\end{table}

\begin{table*}[t]
    \small
    \setlength{\tabcolsep}{4.9pt}
    \resizebox{1\linewidth}{!}{
	\centering
	\begin{tabular}{l|l|cccc|cccc}
	\toprule
        \multirow{2}{*}{\textbf{Methods}}&\multirow{2}{*}{\textbf{Backbone}}&\multicolumn{4}{c}{\textbf{\taskone{}}} &\multicolumn{4}{c}{\textbf{\tasktwo{}}}\\
        \cmidrule(lr){3-6}\cmidrule(lr){7-10}
	&&\textbf{Easy}&\textbf{Medium} &\textbf{Hard} &\textbf{Avg.}&\textbf{Easy}&\textbf{Medium} &\textbf{Hard} &\textbf{Avg.} \\
            \midrule
            \textbf{Random} &  - & 25.00 & 25.00 & 25.00 & 25.00 & 25.00 & 25.00 & 25.00 & 25.00 \\
            \textbf{Majority Vote} & - & 26.37	& 25.24	& 26.27	& 26.15	& 25.97	& 28.57 & 28.57 & 26.60 \\
            \midrule
            \multirow{5}{*}{\textbf{PTLM}}&RoBERTa-Large \scriptsize{\textit{214M}} &41.46&41.98&38.98&41.43&54.95&35.06&30.08&49.84\\
            & DeBERTa-v3-Large \scriptsize{\textit{435M}}&36.40&38.72&37.62&36.90&26.52&29.35&32.33&27.39\\
            &T5-v1.1-xxl \scriptsize{\textit{11B}}&24.84&25.47&25.42&24.99&26.71&26.23&25.56&26.55\\
            &Flan-T5-xxl \scriptsize{\textit{11B}}&75.98&73.58&63.56&74.88&79.26&81.82&81.95&79.89\\
            &T0-pp \scriptsize{\textit{11B}}&71.70&68.87&64.41&70.78&77.11&76.10&78.20&76.99\\
		  \midrule
            \multirow{5}{*}{\textbf{Commonsense}}&HyKAS \scriptsize{\textit{435M}}&71.81&67.17&46.69&69.61&47.02&45.97&48.12&46.90\\
            &CAR \scriptsize{\textit{435M}}&73.69&71.46&54.38&72.20&36.18&43.12&44.36&37.94\\
            &CANDLE \scriptsize{\textit{435M}}&74.34&70.75&52.54&72.52&35.94&43.90&43.61&37.84\\
            &VERA \scriptsize{\textit{11B}}&69.82&70.52&61.02&69.49&59.20&58.18&64.66&59.36\\
            &VERA-CANDLE \scriptsize{\textit{11B}}&70.59&71.33&63.41&70.02&62.18&60.13&66.13&61.81\\
            \midrule
            \multirow{8}{*}{\textbf{Open LLM}}&LLAMA2-7B-chat&64.98&66.54&53.85&64.61&59.90&54.86&47.37&58.04\\
            &LLAMA2-13B-chat&69.63&63.96&60.78&68.06&45.53&41.95&39.71&44.52\\
            &Gemma-2B-instruct&48.77&47.23&48.21&48.45&39.45&39.15&38.17&39.32\\
            &Gemma-7B-instruct&65.55&64.31&52.04&64.61&33.18&36.01&41.51&34.20\\
            &Mistral-7B-Instruct-v0.2&76.57&74.53&63.56&75.50&59.78&62.60&65.41&60.64\\
            &Falcon-7B-instruct&24.54&22.17&28.26&24.25&26.15&28.05&26.32&26.50\\
            &Vicuna-7B-v1.5&57.13&57.08&55.43&57.05&27.88&30.13&23.31&28.00\\
            &LLAMA3-70B&70.88&64.68&61.87&68.71&46.32&43.18&40.71&44.91\\
            &LLAMA3.1-70B&72.31&65.72&62.18&69.98&48.57&44.03&41.87&46.35\\
            \midrule
            \multirow{2}{*}{\textbf{\name{} Distilled}}&LLAMA2-7B-chat&65.78&64.61&55.75&66.15&59.43&57.13&60.03&59.04\\
            &Mistral-7B-Instruct-v0.2&78.57&74.31&80.89&76.97&61.14&65.42&62.16&62.02\\
            \midrule
            \multirow{6}{*}{\textbf{LLM API}}
            &ChatGPT 
            &75.06&73.76&68.64&74.48&80.74&76.62&68.42&79.23\\
            &ChatGPT (CoT) &76.07&74.53&63.56&75.12&78.89&75.32&78.20&78.21\\
            &ChatGPT (CoT-SC) &76.51&73.82&63.56&75.32&85.72&77.14&82.71&83.99\\
            &GPT 4 &78.12&75.41&66.10&76.97&86.03&82.34&84.96&85.30\\
            &GPT 4 (CoT) &78.12&75.41&66.10&76.97&86.03&82.34&84.96&85.30\\
            &GPT 4 (CoT-SC) &78.80&72.88&65.25&76.97&84.00&80.78&84.96&83.48\\
            \midrule
            \midrule
            \textbf{Human} & - &89.96&90.00&80.96&89.33&95.50&85.19&100.0&94.00\\
		\bottomrule
	\end{tabular}
 }
  \vspace{-0.1in}
	\caption{Evaluation results (Accuracy\%) of various language models on both tasks of the IntentionQA benchmark.}
 \vspace{-0.2in}
    \label{tab:main_eval_results}
\end{table*}

\section{Experiments and Analyses}
In this section, we first study the downstream benefits brought by intentions generated by~\name{}.
Then, we conduct in-depth analyses to demonstrate the advantages of multimodal generation in~\name{} compared to generating only with textual information, the superior capability of the human-centric role-aware filter in comparison to other filtering measures, knowledge diversity in~\name{} generations, and its robustness when generating with different prompts.

\subsection{Evaluation Setup}
\label{sec:experiment_setup}
We explore the effectiveness of \name{} on the IntentionQA benchmark~\cite{ding2024INTENTIONQA}, a comprehensive multiple-choice question answering dataset comprising two challenging subtasks that require language models to comprehend and utilize intentions in E-commerce scenarios accurately.
The first task assesses LLMs' capability in accurately inferring the intention given a co-buy product pair together with 3 distractors sampled from other product pairs, while the second task evaluates LLMs' capability in utilizing purchase intention to make reasonable product recommendation by selecting the product that best aligns with the user's intention from four choices.

While existing results show that language models struggle with both tasks, we aim to examine whether \name{} can enhance LLMs' intention understanding capabilities through fine-tuning.
Specifically, from all intentions generated by~\name{}, we transform them into instruction-following format via natural language templates following~\citet{DBLP:journals/corr/abs-2311-07911}.
Then, we fine-tune LLAMA2-7B-chat~\cite{DBLP:journals/corr/abs-2310-06825} and Mistral-7B-Instruct-v0.2~\cite{DBLP:journals/corr/abs-2307-09288} on the retrieved data as a type of knowledge injection.
Specifically, we adopt LLaMA-Factory~\cite{DBLP:journals/corr/abs-2403-13372} through our fine-tune process, maintaining default hyperparameters and a LoRA rank of 64. The fine-tune instruction is attached in Appendix~\ref{appendix:Fine-tune Instructions}.
All experiments are performed on a Linux machine with eight NVIDIA V100 GPUs.
They are then evaluated in a zero-shot manner by being prompted to select the most plausible choice for every QA pair in IntentionQA.
The zero-shot evaluation setup could be found at Appendix~\ref{appendix:Zero-shot Setup}.
Accuracy is used as the evaluation metric. 

\subsection{Baseline Backbone}
\label{appendix:Baseline_Backbone}
For both tasks, we first incorporate random and majority voting to reflect the characteristics of the benchmark.
Five Pre-Trained Language Models (PTLMs) are included: RoBERTa~\cite{DBLP:journals/corr/abs-1907-11692}
DeBERTa-v3~\cite{DBLP:conf/iclr/HeGC23}, T0~\cite{DBLP:conf/iclr/SanhWRBSACSRDBX22}, T5~\cite{DBLP:journals/jmlr/RaffelSRLNMZLL20}, and Flan-T5~\cite{DBLP:journals/corr/abs-2210-11416}.
Then, performances by five commonsense-injected PTLMs are also reported, including HyKAS~\cite{DBLP:conf/aaai/MaIFBNO21}, CAR~\cite{DBLP:conf/emnlp/WangF0XLSB23}, VERA~\cite{DBLP:conf/emnlp/0010WWS0H23}, CANDLE~\cite{DBLP:journals/corr/abs-2401-07286}, and VERA-CANDLE. 
We also report the performances of several LLMs, such as LLaMA2~\cite{DBLP:journals/corr/abs-2307-09288}, Gemma~\cite{DBLP:journals/corr/abs-2403-08295}, Mistral~\cite{DBLP:journals/corr/abs-2310-06825}, ChatGPT~\cite{openai2022chatgpt}, and GPT-4~\cite{GPT4}.
For the latter two, we also adopt Chain-of-Thought (\textsc{CoT};~\citealp{wei2022chain}) and CoT with Self-Consistency (\textsc{CoT-SC};~\citealp{wang2022self}) prompting.



\subsection{Results}
The results are presented in Table~\ref{tab:main_eval_results}, demonstrating significant improvements in both tasks when LLMs are fine-tuned on intentions generated by~\name{}.
For instance, LLAMA2 achieves accuracy gains of 1.54\% and 1.00\% for both tasks, respectively.
Notably, Mistral yields a remarkable performance gain that even becomes comparable to GPT-4, despite having a significantly lower number of parameters.  
However, for the intention utilization task, while both fine-tuned LLMs show performance improvements, they still fall behind GPT-4. 
One potential reason for this gap could be the misalignment between the fine-tuning objective and the evaluated ability of the task, which involves generating intentions for a pair of products and selecting a product based on a given intention. 
Nevertheless, these results underscore the effectiveness and efficiency of~\name{} in enhancing LLMs' capabilities in E-commerce intention comprehension and utilization.

\subsection{Analyses}
In this section, we study the superiority of~\name{} by examining three aspects. 
First, we demonstrate the positive impact of acquiring intentions in a multimodal manner instead of relying solely on textual hints. 
Next, we show that our proposed human-centric filtering leads to better downstream results and is more effective than traditional critic filtering based on a supervised scoring discriminator. 
Finally, we illustrate the robustness of~\name{} when using different prompts and its superior quality compared to FolkScope.

\subsubsection{Multimodal vs. Unimodal Generation}
We first study the ablation of incorporating visual information in~\name{} by comparing the downstream benefits of intentions generated in multimodal versus unimodal (text-only) paradigms. 
For a fair comparison, we exclude visual input in LLaVa when generating in a unimodal manner and instruct it to generate intentions for the same purchasing records as in~\name{} with prompts that are as identical as possible. 
We then fine-tune LLMs on the collected intentions, evaluate the resulting models on the IntentionQA benchmark, and compare the performance of the two types of distilled models. 
The results are shown in Table~\ref{tab:ablation_eval_result}.
We observe that fine-tuning LLMs on intentions generated with textual information can only merely improve their performances on downstream tasks, certifying the need of additional visual signals.


\begin{figure}
    \centering
    \includegraphics[width=1\linewidth]{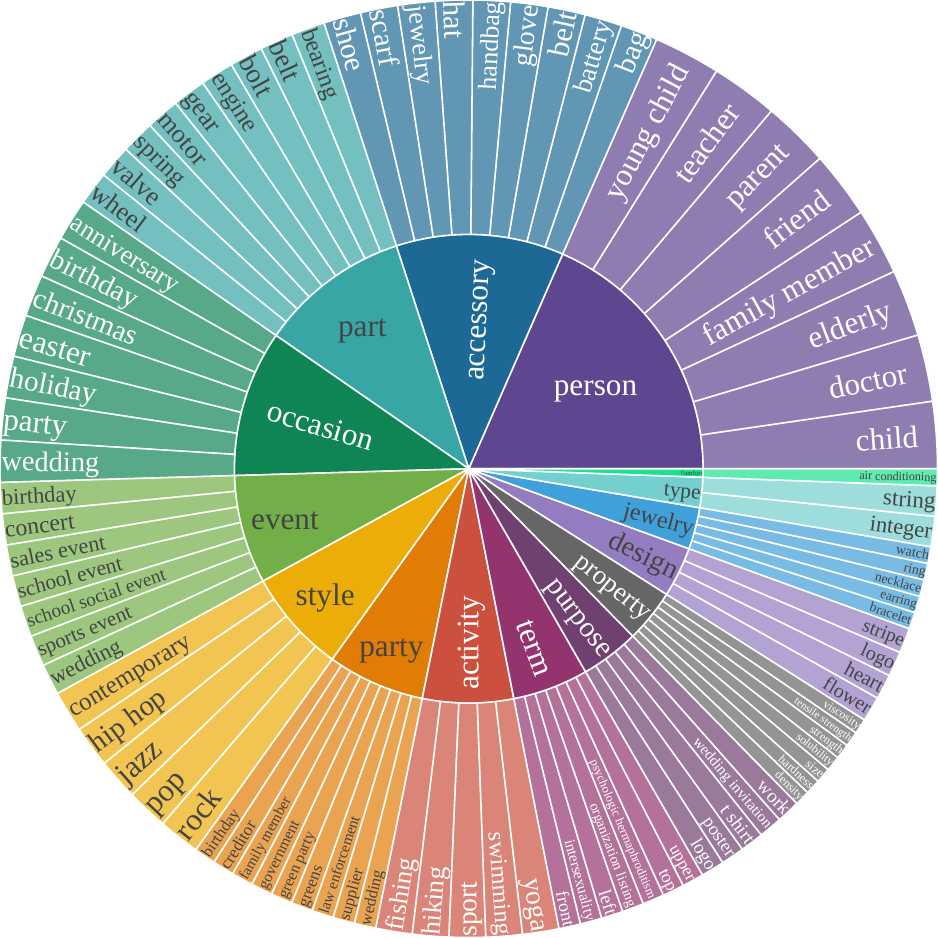}
    \caption{Distribution of hypernyms sourced from Probase in \name{} with top frequencies.}
    \label{fig:diversity_comparison}
    \vspace{-0.2in}
\end{figure}

\subsubsection{Diversity of \name{}}
Moreover, the semantic diversity of the generated intentions are another flag of the quality as featuring intentions that cover a diverse collection of topics, events, and even mental states makes it more possible to model purchase intentions comprehensively.
Thus, following~\citet{DBLP:journals/corr/abs-2401-07286}, we sample 30,000 intentions from \name{}, extract the nouns in them via dependency parsing, and plot the distribution of the hypernyms of these nouns, matched against Probase~\cite{DBLP:conf/sigmod/WuLWZ12}, according to their number of occurences.
The resulting plot is shown in Figure~\ref{fig:diversity_comparison}.
Remarkably, the intentions generated by \name{} display elevated levels of diversity, which signifies the broad semantic coverage of purchasing different products in the generated intentions.
We posit that such high semantic diversity may provide implicit benefits to downstream tasks when employing~\name{} in E-commerce applications.


\subsubsection{Impact of Role-Aware Filter}
\begin{figure}[t]
    \centering
    \includegraphics[width=1\linewidth]{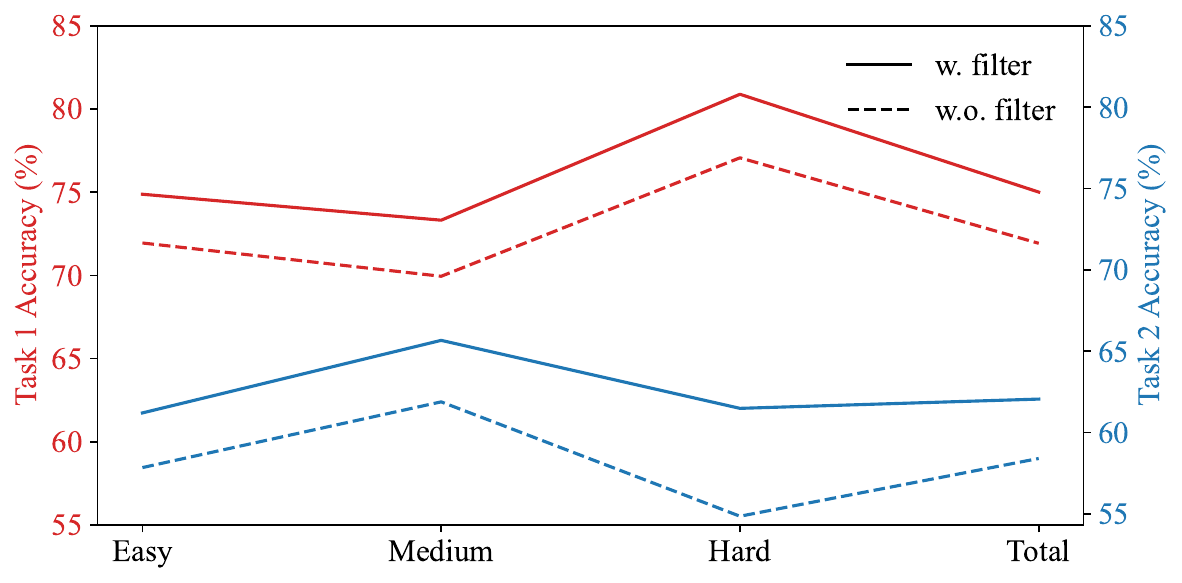}
    \caption{
    Ablation results on IntentionQA tasks by Mistral-7B distilled on intentions generated by~\name{} with/without filtering.
    }
    \label{fig:mind_ablation}
\end{figure}

\paragraph{Ablation Study on IntentionQA.}
We then study the ablation of~\name{} by focusing on the role of our proposed human-centric role-aware filter mechanism in its impact toward quality control of the generated intentions.
Specifically, we leverage the IntentionQA~\cite{ding2024INTENTIONQA} as the evaluation benchmark and separately train two models on (1) intentions that are filtered by our proposed mechanism (\textit{w. filter}) and (2) intentions without filtering (\textit{w.o. filter}).
All setups follow the same as described in Section~\ref{sec:experiment_setup}, and we use Mistral-instruct-7B-v0.2 as the backbone and train it using a unified hyper-parameter setting.
The results are plotted in Figure~\ref{fig:mind_ablation}.
We observe that, without filtering, performances on both tasks across all difficulty levels drop significantly, which is possibly due to the inclusion of more noisy intentions in the training data.
This shows that our proposed filtering module is indeed functioning well in controlling high-quality intentions and is beneficial to downstream tasks.

\begin{table*}[t]
    \small
    \centering

    \resizebox{0.95\linewidth}{!}{
	\centering
	\begin{tabular}{l|l|cccc|cccc}
	\toprule
        \multirow{2}{*}{\textbf{Backbones}}&\multirow{2}{*}{\textbf{Training Recipe}}&\multicolumn{4}{c}{\textbf{\taskone{}}} &\multicolumn{4}{c}{\textbf{\tasktwo{}}}\\
        \cmidrule(lr){3-6}\cmidrule(lr){7-10}
	&&\textbf{Easy}&\textbf{Medium} &\textbf{Hard} &\textbf{Avg.}&\textbf{Easy}&\textbf{Medium}          &\textbf{Hard} &\textbf{Avg.} \\
        \midrule
        \multirow{4}{*}{\textbf{LLAMA-7B-chat}}& Zero-shot &64.98&\textbf{66.54}&53.85&64.61&\textbf{59.90}&54.86&47.37&58.04\\
        &w. Unimodal & 65.03 & 65.49 & \textbf{56.71} & 64.99 & 59.08 & 54.71 & 45.59 & 57.34 \\
        &w. Critic Filter & 61.88 & 64.56 & 51.22 & 61.67 & 59.27 & 54.13 & 46.89 & 57.88 \\
        &\textbf{MIND Distilled} &\textbf{65.78}&64.61&55.75&\textbf{66.15}&59.43&\textbf{57.13}&\textbf{60.03}&\textbf{59.04} \\
        \midrule
        \multirow{4}{*}{\textbf{Mistral-7B-Instruct-v0.2}}& Zero-shot &76.57&\textbf{74.53}&63.56&75.50&59.78&62.60&\textbf{65.41}&60.64\\
        &w. Unimodal & 75.02 & 72.33 & 62.17 & 73.72 & 58.35 & 61.48 & 62.81 & 58.51\\
        &w. Critic Filter & 74.78 & 71.23 & 62.87 & 72.29 & 58.32 & 61.09 & 58.98 & 57.63\\
        &\textbf{MIND Distilled} &\textbf{78.57}&74.31&\textbf{80.89}&\textbf{76.97}&\textbf{61.14}&\textbf{65.42}&62.16&\textbf{62.02}\\
		\bottomrule
	\end{tabular}
        }
	\caption{Ablation experiment results (Accuracy\%)  on IntentionQA benchmark.}
    \label{tab:ablation_eval_result}
    \vspace{-0.2in}
\end{table*}

\paragraph{Critic Filter v.s. Role-Aware Filter.}
Afterward, we validate the effectiveness of our role-aware filter by comparing it against a traditional critic filter provided by~\citet{DBLP:conf/acl/YuWLBSLG0Y23}. 
The critic filter is obtained by training a language model with a regression objective to predict the typicality of generated intentions in the range of 0 to 1. 
We adopt the released critic scorer, pre-trained on annotated intentions in FolkScope, and use it to score intentions in~\name{} under identical settings. 
By setting a critic threshold to 0.8 and discarding intentions below this threshold, we obtain a sibling subset of~\name{}. 
LLMs are then fine-tuned on this sibling subset and evaluated on the testing sets of IntentionQA. 
The results are presented in Table~\ref{tab:ablation_eval_result}.
It can be observed that LLMs exhibit inferior performance when using the critic filtering mechanism. 
One possible reason is that the pre-trained critic filter only captures the pattern of intentions at different levels of typicality without considering their relation to the products. 
This further verifies the need for a role-aware filtering mechanism.

\subsubsection{Robustness of \name{}}
According to~\citet{DBLP:journals/corr/abs-2404-01077}, generations by LLMs can be significantly impacted by even slight changes in the prompts.
This warrants a potential weakness of~\name{} which heavily relies on prompting in collecting intentions.
Hence, we aim to overcome this by proving that intentions generated with modified prompts are generally semantically consistent at high quality.
Specifically, we exclude the prompts which explicitly instructing the LVLMs to rely on visual cues from the product images and only retain the prompts that require the LVLMs to generate intentions.
Then, 100 product pairs are randomly sampled from~\name{} to generate intentions utilizing the modified prompts.
Finally, the sentence embedding are calculated using SentenceBERT~\cite{DBLP:conf/emnlp/ReimersG19}, and the cosine similarity between each modified intention and its corresponding original intention generated by \name{} is derived.
The results revealed an average cosine similarity of 0.85 between the intentions generated with modified prompts and those generated by \name{}. This high similarity indicates the robustness of intention generation process.
Inter-relation intention comparison examples are provide in Appendix~\ref{appendix:inter_relation}

\begin{figure}[t]
    \centering
    \includegraphics[width=0.95\linewidth]{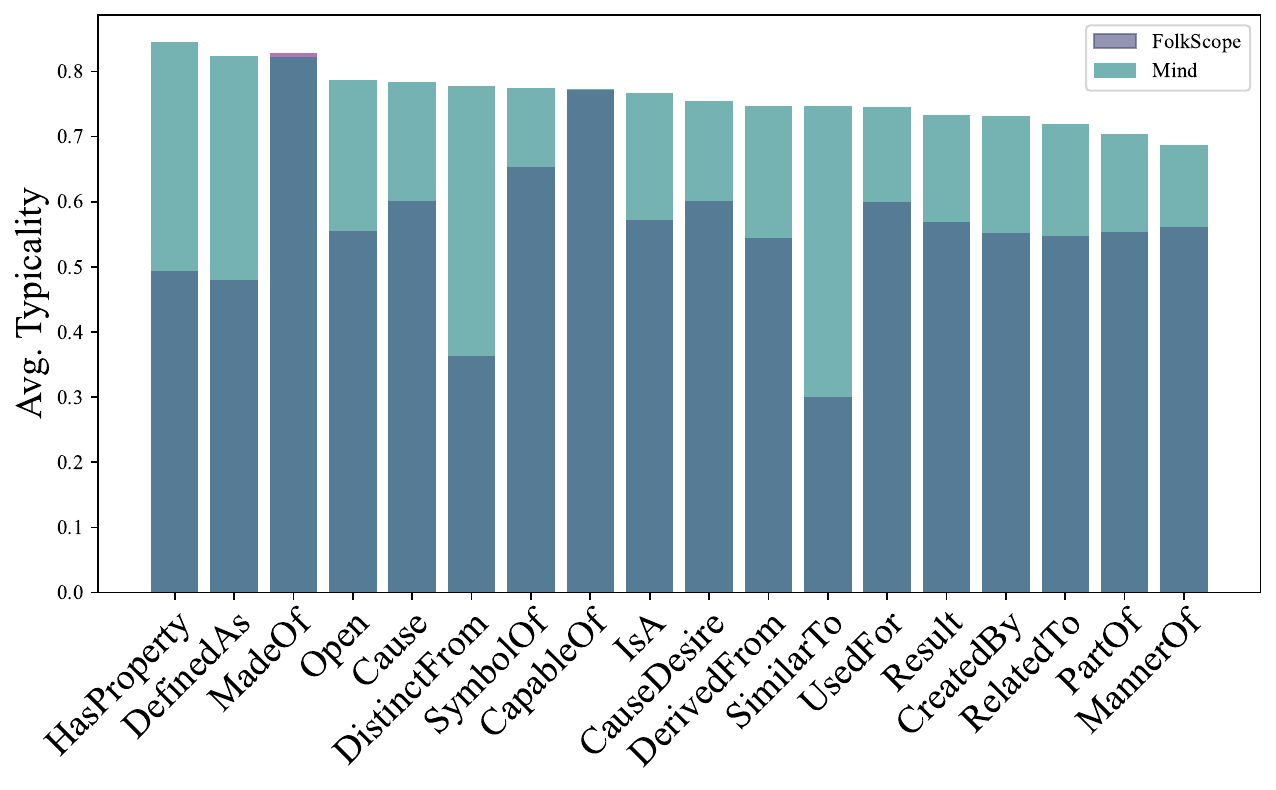}
    \caption{
    Relation-wise comparison of typicality scores across all relations between~\name{} and FolkScope.
    }
    \label{fig:mind_vs_fkscp}
    \vspace{-0.2in}
\end{figure}

\subsubsection{Comparisons Against FolkScope}
We then compare~\name{} against FolkScope, the previous state-of-the-art method for large-scale intention acquisition, by analyzing the typicality distribution of intentions across all relations.
Specifically, we adopt the same annotation protocols designed by~\citet{DBLP:conf/acl/YuWLBSLG0Y23,MARS,CAT,ConceptSurvey} and transfer our annotation results into a four-point Likert scale~\cite{joshi2015likert}.
Then, for each relation, we compute the average typicality scores among all intentions and plot them for comparison, as shown in Figure~\ref{fig:mind_vs_fkscp}.
From the plot, we observe that intentions generated by~\name{} exhibit higher typicality scores across nearly all relations compared to those generated by FolkScope, which further demonstrates the superiority of~\name{}.
More relation-wise filter result analysis is attached in Appendix~\ref{appendix:relation-wise-filter}.

\begin{table*}[t]
\small
\centering
{\def\arraystretch{1}
\begin{tabularx}{\textwidth}{>{\raggedright}p{4cm}|>{\RaggedRight\arraybackslash}p{4cm}|p{6.5cm}}
\toprule
\centering Item1 & \centering Item2 & Intentions \\ \midrule
Samsung SmartCam HD Pro &Samsung SmartThings Smart Home Hub & \textcolor{mind_color}{They are designed to \textbf{work together in a smart home system}}\\  
& &\textcolor{fkscp_color}{They are derived \textbf{from the same category.}}\\ 
\midrule
Clarks Women's Ankle Bootie & The Sak Kendra Hobo Shoulder Bag & \textcolor{mind_color}{The consumer is looking for \textbf{a stylish and functional combination} for their daily activities.}\\
& &\textcolor{fkscp_color}{They both are \textbf{a manner of 'Women's Shoes' and 'Women's Handbags'} respectively.}\\ 
\midrule
Western Party Mustaches & \multirow{2}{*}{\shortstack{Forum Novelties Adult\\ Cowboy Costume Vest}} & \textcolor{mind_color}{They are both part of \textbf{a costume or a themed party.}}\\& &\textcolor{fkscp_color}{They both are a \textbf{part of the 'Adult Costume' category.}}\\ 
\midrule
Columbia Women's Loveland Shorty Omni-Heat Snow Boot &Columbia Sportswear Women's Thermarator Glove & \textcolor{mind_color}{They are designed to \textbf{keep the wearer warm and comfortable during cold weather conditions}}\\& &\textcolor{fkscp_color}{They both are a \textbf{part of the Columbia brand.}}\\ 
\midrule
Banded Arc Welded Waterproof Backpack Polyester&Banded Deluxe UFS Fleece Face Mask & \textcolor{mind_color}{They are both used for \textbf{outdoor activities and protection from harsh weather conditions.}}\\
& & \textcolor{fkscp_color}{They are both used for \textbf{outdoor activities.}}\\
\bottomrule
\end{tabularx}
}
\caption{Case studies of purchase intentions generated by \textcolor{mind_color}{\name{}} and \textcolor{fkscp_color}{FolkScope}. Intentions generated by~\name{} are highlighted in \textcolor{mind_color}{\textbf{purple}} and those generated by FolkScope are marked in \textcolor{fkscp_color}{\textbf{orange}}.}
\vspace{-0.2in}
\label{tab:case_study}
\end{table*}

\subsubsection{\name{} Against FolkScope Case Study}
Aside from empirical analyses, we also show the advantages of~\name{} over FolkScope through additional case studies to highlight key benefits of~\name{}. 
To this end, we randomly selected 5 pairs of co-buy products and compared the intentions generated by both frameworks, as shown in Table~\ref{tab:case_study}. 
Our findings from the table indicate that~\name{}-generated intentions exhibit a stronger focus on the usage and functionalities that potentially fulfill customers' needs and intentions when purchasing these products. 
Conversely, intentions generated by FolkScope tend to be biased towards properties and features that can be easily inferred from the product titles, which are of lesser interest to customers' shopping intentions.
Take the second row in Table~\ref{tab:case_study} as an example. 
The intentions \textcolor{fkscp_color}{\textit{both are ``Women's Shoes'' and ``Women's Handbags''}} generated by FolkScope merely represent an aggregation of the product categories for the two items. 
In contrast,~\name{} produces intentions such as \textcolor{mind_color}{\textit{looking for stylish and functional combination for daily activities}} , which better captures a customer's intention when shopping for both products. 
This example further reinforces our previous conclusions that~\name{} can generate intentions that are more human-centric and better reflect the customers' intentions as mental states.

\vspace{0.03in}
\section{Conclusions}
\vspace{0.03in}
In this work, we present \name{}, a multimodal distillation framework for enhancing E-commerce purchase understanding by automating the pipeline of intention generation and quality filtering via multiple-step instructions over LVLMs.
By applying~\name{} to real-world E-commerce data, we construct the very first multimodal purchase intention knowledge base featuring over 1.2 million intentions.
These intentions have been proven to be invaluable in distilling student models that exhibit improved performance in E-commerce intention comprehension and utilization tasks.
Further analyses reveal the effectiveness of~\name{} by validating the proposed filtering mechanism and highlighting the strengths of~\name{} in comparison to FolkScope.
Our work sheds light on improving large-scale E-commerce intention acquisition and application.

\clearpage
\section*{Limitations}
First, \name{} generates intention by leveraging several zero-shot prompts without additional exemplars. 
This decision is made as we observe that few-shot prompts may ``guide'' LVLM to generate intentions that tend to be similar to the provided exemplars, which harms diversity. 
However, it remains an open question whether more advanced prompting methods~\cite{DBLP:journals/csur/Song0CMS23, DBLP:journals/corr/abs-2203-04291} would help in the generation process. 
It's also worth noting that the LVLM used in our work may be outdated as new products show up on E-commerce stores. 
However, switching LLaVa to more up-to-date LVLMs, preferably pre-trained on E-commerce data, can address this concern. 
Finally, \name{} utilizes an automatically functioning filter as quality control. 
While we have shown its effectiveness, it remains challenging to effectively regulate the filter mechanism to be either lenient or strict. 
Further investigation is required to provide insights into the alignment between the values of VLMs and the real world, enhancing our understanding of them.

\section*{Ethics Statement}
To avoid generating harmful intentions and toxic filter rationales in \name{}, we recruit 4 expert annotators who are graduate students specializing in multilmodality and natural language processing to evaluate the generated intentions and rationales. 
We ask all experts to go through 200 sampled data and no harmful contents are reported.
The crowd-sourced annotators are paid a wage that complies with the local law.
The expert annotators involved in this research are knowledgeable about the annotation protocol and the intended utilization of their annotations. 
They are willingly to contribute without expecting any compensation.
The training and evaluation datasets utilized in this study are publicly available, anonymized, and shared under open-access licenses for research purposes, adhering to their intended usage. 
Thus, we believe this paper does not yield any ethical issue.

\section{Acknowledgements}
We thank the anonymous reviewers and the area chair for their constructive comments.
The authors of this paper were supported by the NSFC Fund (U20B2053) from the NSFC of China, the RIF (R6020-19 and R6021-20), and the GRF (16211520 and 16205322) from RGC of Hong Kong. 
We also thank the support from Amazon.



\bibliography{anthology,custom}

\appendix

\begin{center}
    {\Large\textbf{Appendices}}
\end{center}

\section{Prompts}
\label{appendix:prompts_used}
In this section we show the instructions used in feature extraction, intention generation, and human-centric role-aware filtering stages. 
The prompts are shown in Table~\ref{tab:prompts_used}.

\begin{table*}[t]
\centering
\small
\begin{tabularx}{\textwidth}{>{\raggedright}p{5cm}|>{\RaggedRight\arraybackslash}p{9.5cm}}
\toprule
Task & Prompt \\ 
\midrule
\multirow{2}{*}{Feature Extraction} & 
 The $[IMAGE\_1,IMAGE\_2]$ contains a product and name of it is $[PROD\_NAME]$. Please analyze the product image, together with the product name, provide a detailed description focusing on the product's features, design, and apparent quality. Highlight any unique characteristics or visible elements that distinguish this product from similar items. Additionally, speculate on the potential uses and benefits of this product for a consumer, based on its appearance or any information in the image and the name.\\
\midrule
Intention Generation & The two $[Image\_1,Image\_2]$ are two different products. The product name of the upper image is $[Prod\_A\_Name]$. The product detail and the potential purchase intention is ${Prod\_A\_Desc}$. The product name of the lower image is ${Prod\_B\_Name}$. The product detail and the potential purchase intention is ${Prod\_B\_Desc}$. Based on information provided, together with the product images, what could be the potential intention for people buying these two products in one purchase simultaneously based on the relation of $[Relation\_Prompt[Relation]$, take the image features into consideration, limit your word count within 120 words. Start with the potential co-buy intention could be ${Relation\_Prompt[Relation]}$\\ 
\midrule
Human-centric Role-aware Filtering & The two images $[Image\_1,Image\_2$ are two different products. The product name of the upper image is $[Prod\_A\_Name$. The product detail and the potential purchase intention is $[Prod\_A\_Desc]$. The product name of the lower image is $[Prod\_B\_Name$. The product detail and the potential purchase intention is $[Prod\_B\_Name]$. Under the relation of $[Relation\_Prompt[Relation]$, the potential co-buy intention would be $[Intention]$. If you are a consumer who are eager to buy product a or product b, would this intention encourage you to buy the two products simultaneously? be critical on your choice, output yes or no together with the reason for your answer. For example, the output should be Yes, ... or No, ... \\  
\bottomrule
\end{tabularx}
\caption{Prompts used for evaluating LLM baselines across various tasks in a zero-shot scenario.}
\label{tab:prompts_used}
\end{table*}

\section{Annotator Requirement}
\label{appendix:annotator_requirement}
For strict quality control, we only invite workers satisfying the following requirements: 1) at least 1K HITs approved, and 2) at least 95\% approval rate.
Then, we conduct two rounds of qualification rounds using a qualification question set crafted by authors of this paper, which includes both straightforward and tricky questions.
Over 600 workers participated and only 90 (15\%) of them are deemed qualified by achieving over 87\% accuracy.

\section{Error Analysis of Filtered Intentions}
\label{appendix:error_analysis_filter}
While human annotation results in Section~\ref{sec:annotation_results} show that, after filtering, most of the remaining intentions are highly plausible and typical, we observe that only 46.7\% generations passed our proposed filtering module as the last step of~\name{}.
Thus, in this section, we first study the role of such human-centric filtering by looking into the causes of why the intentions get discarded, and further seek insights to resolve such a high filtering loss.
To achieve this, we randomly sample 200 intentions that are abandoned by~\name{} during the last step and manually annotate the reasons behind based on the rationale provided by the LVLM.
Three types of errors are observed and they are categorized as:
\begin{itemize}[leftmargin=*]
    \item 81.0\% of the filtered intentions, while plausible, do not provide strong enough evidence to motivate a LVLM agent to execute the purchase behavior for two products.
    For example, the intention ``\textit{they both are related to home audio systems}'' for purchasing a pair of audio adapters lacks customer interaction and solely focuses on the products themselves.
    A more appropriate intention, for example, ``\textit{they both are able to help in connecting audio devices},'' would retain a stronger bond between the products and customers by aligning with their functionalities.
    \item 13.0\% of the intentions result from misjudgment by the LVLM, where the agent fails to make the correct decision despite the intention being sufficiently plausible and typical. 
    This highlights the need for future improvements, including a more refined filter to enhance our framework.
    \item 6.0\% of the intentions are discarded due to being implausible or containing factual errors that do not align with the products.
\end{itemize}
Overall, 87\% of intentions are being properly discarded, which is considerably high for an automatic filter without human supervision.

\section{Fine-tuning Instructions}
\label{appendix:Fine-tune Instructions}
The instruction adopted during the fine-tuning process is attached below:
\begin{tcolorbox}[title={Prompt Template for Fine-tuning}, colback = cBlue_1!10, colframe = cBlue_7,  coltitle=white,fonttitle=\bfseries\small, center title,fontupper=\small,fontlower=\small]
\textbf{Question}: Q: customer buys <product 1> and <product 2>. What is the most likely intention for buying them?
\tcblower
\textbf{Answer}:<intention>.
\end{tcolorbox}
Where <product 1> and <product 2> are the products co-bought together and the <intention> refers to the co-buy intention retreived from \name{}.

\section{Zero-shot Evaluation Setup}
\label{appendix:Zero-shot Setup}
For the zero-shot evaluation process, we adopt the same hyperparameter setting as the default model provided bu HuggingFace.
The Zero-shot evaluation prompt are as below:
\begin{tcolorbox}[title={Evaluation Prompt for Intention Understanding}, colback = cBlue_1!10, colframe = cBlue_7,  coltitle=white,fonttitle=\bfseries\small, center title,fontupper=\small,fontlower=\small]
\textbf{Question}: A customer buys <product 1> and <product 2>. What is the \textbf{most likely intention} for buying them?
\tcblower
\textbf{Candidate Answers}: \\
A. because <intention 1>. \\
B. because <intention 2>. \\
C. because <intention 3>. \\
D. because <intention 4>. \\
Answer A or B or C or D only without any other word.
\end{tcolorbox}
The <product 1> and <product 2> are products co-bought together and the 4 intentions in the candidate answers are four intentions originate from FolkScope. According to the restriction of \textbf{most likely}, only one intention could be the correct answer.

\begin{tcolorbox}[title={Evaluation Prompt for Intention Utilization}, colback = cBlue_1!10, colframe = cBlue_7,  coltitle=white,fonttitle=\bfseries\small, center title,fontupper=\small,fontlower=\small]
\textbf{Question}: A customer buys <product>, because <intention>. What is the customer's \textbf{most probable} additional purchase?
\tcblower
\textbf{Candidate Answers}:\\
A. <product 1>\\
B. <product 2>\\
C. <product 3>\\
D. <product 4>\\
Answer A or B or C or D only without any other word.
\end{tcolorbox}
The intention refers to the co-buy intention retreived from FolkScope. The 4 products are the products recorded in FolkScope. According to the restriction of \textbf{most probable}, only one product could be the correct answer.

\section{\name{} Inter-relation Case Study}
\label{appendix:inter_relation}
\begin{figure*}
    \centering
    \includegraphics[width=1\linewidth]{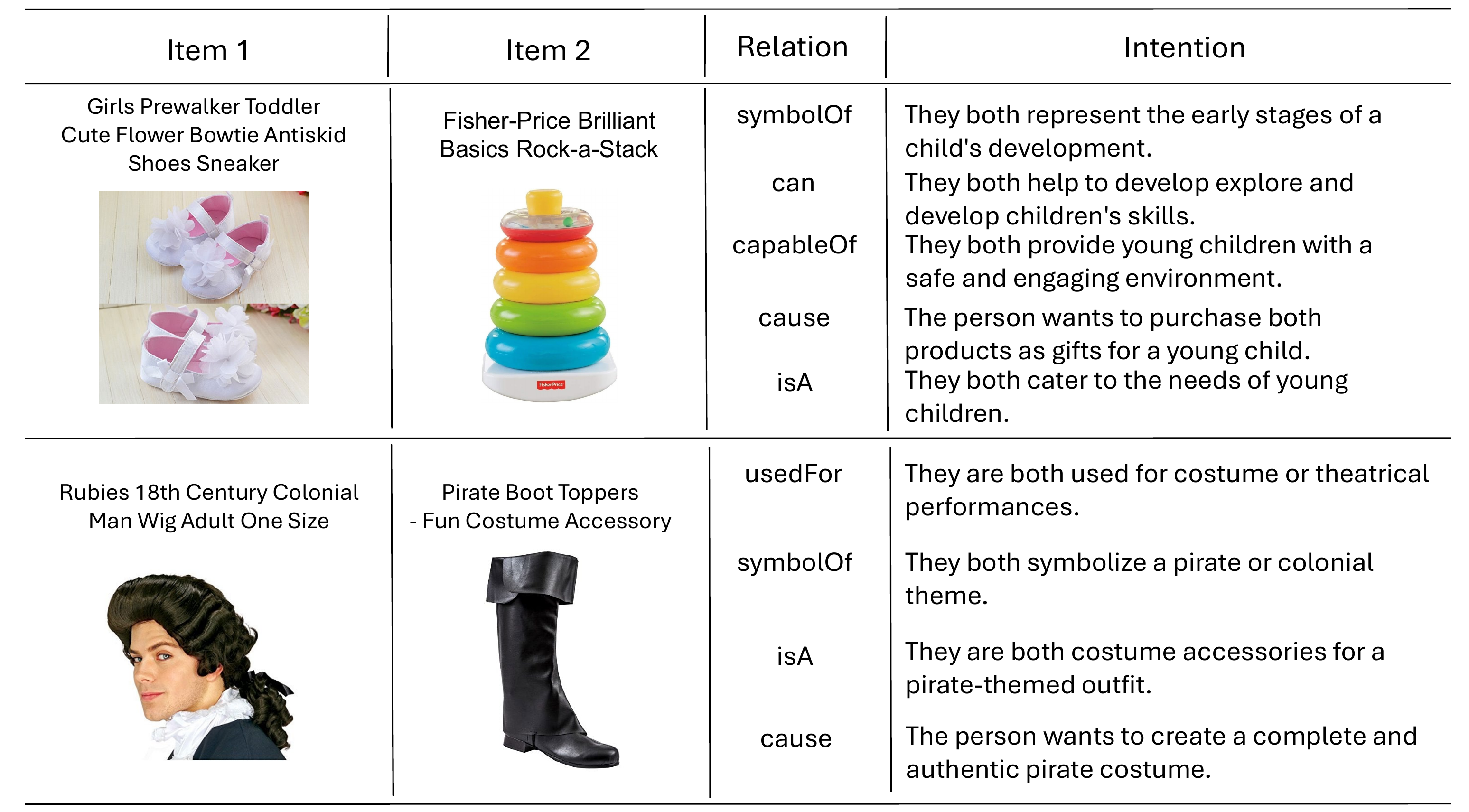}
    \caption{\name{} co-buy intentions generated under different relations.}
    \label{fig:case_study}
\end{figure*}
In this section, we showcase various co-buy intentions for the same product pairs generated under different relations. 
The examples are provided in Figure~\ref{fig:case_study}.

It is evident from the table that the intentions consistently capture the key aspect of the co-buy intention. i.e., for young kids, for costume, pirate.
Though for certain relations the intention doesn't follow the instruction strictly in terms of format, the quality of the intention remains reasonable and informative. 
The content of these intentions is still aligned with the intended purpose of the designed relation.

\section{Relation-wise Filter Analysis}
\label{appendix:relation-wise-filter}
\begin{figure}[t]
    \centering
    \includegraphics[width=0.5\textwidth]{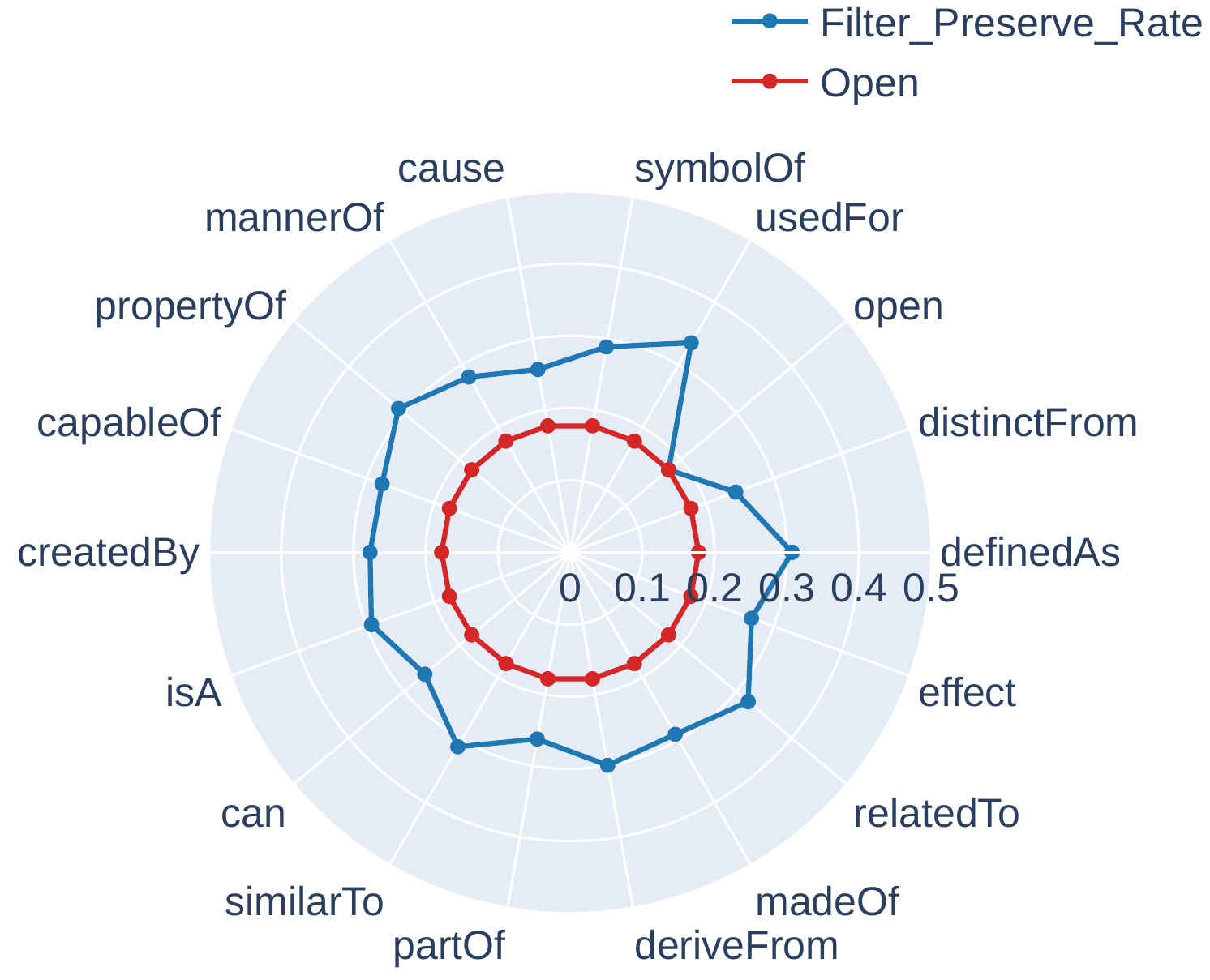}
    \caption{The rate of preserved intentions after filtering under different relations.}
    \label{fig:relation-wise-filter}
\end{figure}

In this section, we present the Relation-wise Filter Preserve Rate (RFP Rate) of \name{}, which represents the proportion of intentions that are retained among all intentions for every relation. 
We report our result in Figure~\ref{fig:relation-wise-filter}.

Our observations indicate that the open relation has the lowest RFP Rate at 0.17 yet other relations demonstrate RFP Rates ranging from 0.2 to 0.4.

We hypothesize that the under-performance of open relation generation could be attributed to its less specific instruction adopted in generation process.
The lack of specific information in the instruction may hinder the LVLM's ability to generate persuasive and informative intentions for the filter LVLM, resulting in the low preserve rate.

This finding emphasizes the importance of future intention mining research. 
It suggests that solely relying on the expressive power of LVLMs to undermine potential intentions is not feasible. 
Instead, a meticulous instruction constraint aligns with research purpose is required. 
Specifically, incorporating detailed relation information during intention mining is indispensable in E-commerce co-buy behavior understanding domain. 
This could improve the intention mining process, leading to a better construction of a credible and comprehensive intention knowledge base.

\end{document}